%% file: ijcai19.tex
\documentclass{article}
\usepackage{ijcai19}

\usepackage{times}
\usepackage{soul}
\usepackage{url}
\usepackage[hidelinks]{hyperref}
\usepackage[utf8]{inputenc}
\usepackage[small]{caption}
\usepackage{graphicx}
\usepackage{amsmath}
\usepackage{booktabs}
\usepackage{algorithm}
\usepackage{multirow}
\usepackage[dvipsnames]{xcolor}
\usepackage{tikz}
\usetikzlibrary{matrix}
\tikzset{
  matrixstyle/.style={
    matrix of nodes,
    nodes in empty cells,
    column sep      = -\pgflinewidth,
    row sep         = -\pgflinewidth,
    nodes={%
      inner sep=0mm,outer sep=0pt,
      minimum size=12mm,
      text height=\ht\strutbox,text depth=\dp\strutbox,
      draw
    },
    ampersand replacement=\&
}}

\usepackage{amssymb}
\usepackage{amsthm}
\usepackage[noend]{algpseudocode}
\usepackage{enumitem}

\newtheorem{definition}{Definition}
\newtheorem{proposition}{Proposition}

\newcommand{\stateSpace}{\mathcal{S}}
\newcommand{\actionSpace}{\mathcal{A}}

\newcommand{\transitionFunction}{T}
\newcommand{\rewardFunction}{R}
\newcommand{\corruption}{C}
\newcommand{\tuple}[1]{\left\langle #1 \right\rangle}
\newcommand{\rawMDP}{\stateSpace, \actionSpace, \transitionFunction, \rewardFunction}
\newcommand{\MDP}{\tuple{\rawMDP}}
\newcommand{\rawCRMDP}{\rawMDP, \corruption}

\newcommand{\rawSCRMDP}{\rawCRMDP, d, \LV}
\newcommand{\SCRMDP}{\tuple{\rawSCRMDP}}

\newcommand{\absoluteValue}[1]{\left\lvert #1 \right\rvert}
\DeclareMathOperator{\LV}{LV}
\DeclareMathOperator{\TLV}{TLV}
\DeclareMathOperator{\NLV}{NLV}
\DeclareMathOperator*{\E}{\mathbb{E}}
\DeclareMathOperator{\rllb}{lb}
\DeclareMathOperator{\rulb}{ub}

\makeatletter
\newcommand{\printfnsymbol}[1]{%
  \textsuperscript{\@fnsymbol{#1}}%
}
\makeatother

\title{Detecting Spiky Corruption in Markov Decision Processes}

\author{
Jason Mancuso$^1$\thanks{equal contribution}
\and
Tomasz Kisielewski$^2$\printfnsymbol{1} \and
David Lindner$^3$\printfnsymbol{1} \And
Alok Singh$^4$\printfnsymbol{1}
\affiliations
$^1$Dropout Labs\\
$^2$Independent Researcher\\
$^3$ETH Z\"urich\\
$^4$Terrafuse
\emails{
jason@manc.us,
tymorl@gmail.com,
lindnerd@ethz.ch,
alok.singh@berkeley.edu
}}

\begin{document}

\maketitle

\begin{abstract}
Current reinforcement learning methods fail if the reward function is imperfect, i.e. if the agent observes reward different from what it actually receives. We study this problem within the formalism of Corrupt Reward Markov Decision Processes (CRMDPs). We show that if the reward corruption in a CRMDP is sufficiently ``spiky'', the environment is solvable. We fully characterize the regret bound of a Spiky CRMDP, and introduce an algorithm that is able to detect its corrupt states. We show that this algorithm can be used to learn the optimal policy with any common reinforcement learning algorithm. Finally, we investigate our algorithm in a pair of simple gridworld environments, finding that our algorithm can detect the corrupt states and learn the optimal policy despite the corruption.
\end{abstract}

\section{Introduction}

The reward function distinguishes reinforcement learning (RL) from other forms of learning. If the reward function is misspecified, the agent can learn undesired behavior \cite{Everitt2017}. It is an open problem in RL to detect or avoid misspecified rewards \cite{AmodeiOSCSM16}.

\cite{Everitt2017} formalize this problem by introducing \emph{Corrupt Reward Markov Decision Processes} (CRMDPs).
Informally, a CRMDP is a MDP in which the agent receives a
corrupted reward signal. A \emph{No Free Lunch Theorem} for CRMDPs states that
they are unlearnable in general; however, additional assumptions on the problem can lead to
learnable subclasses. In particular, \cite{Everitt2017} introduce a set
of strong assumptions that allow for a quantilizing agent to achieve
sublinear regret. Other assumptions, however, may lead to distinct learnable
subclasses that are also useful in practice.

In this work, we propose a set of assumptions to create such a subclass. Intuitively, our assumptions capture the notion of the corruption being ``spiky'' with respect to a distance measure on the state space.

\subsection{Problem motivation}

CRMDPs naturally capture different notions of reward misspecification such as wireheading, side effects, avoiding
supervision, and sensory malfunction \cite{Everitt2017,AmodeiOSCSM16}. This makes them a useful framework for developing RL agents robust to reward corruption.

Additionally, different approaches to learning reward functions can  be interpreted in the
CRMDP framework, such as semi-supervised RL \cite{finn2016generalizing}, cooperative inverse RL \cite{Hadfield-Menell16}, learning from human preferences \cite{Christiano17}, and
learning values from stories \cite{RiedlH16}. Approaches to learn a reward function from expert input such as inverse reinforcement learning (IRL) \cite{ng2000algorithms} can yield corrupt reward
functions when learning from an expert of bounded rationality or from
sub-optimal demonstrations. CRMDPs may be able to provide new
theoretical guarantees and insights for IRL methods, which are often limited by the assumption that the expert acts nearly optimal.

\subsection{Solution motivation}
Our approach is inspired by connections to work on robustness to noise and fairness in supervised learning. The connection between supervised learning and RL has been discussed before (e.g. see \cite{BartoDiett}); here, we only use it in an informal way to motivate our approach.

In particular, one way to view supervised learning is as a special case of RL. In this interpretation, the RL policy corresponds to the supervised learning model, the actions are to pick a specific label for an input, and the reward is an indicator function that is 1 if the true label matches our pick, and 0 otherwise. The reward is provided by an oracle that can provide the true labels for a fixed training set of samples.

In noisy supervised learning, this oracle can be fallible, meaning the true label of an instance may not match the label the oracle provides. In this setting, the goal is to learn the true reward function despite only having access to a corrupt reward function. It has been observed that deep neural networks can learn in the presence of certain kinds of noise \cite{Rolnick2017,Drory2018}, which suggests that some classes of CRMDPs beyond those investigated in \cite{Everitt2017} can be solved.

For further inspiration, we turn to the field of fairness in supervised classification. \cite{Dwork2012} provide a natural definition of individual fairness using distance metrics on the input and output spaces and a corresponding Lipschitz conditions. Intuitively, a classifier is considered fair if it provides similar labels for similar input samples. Our approach to solving CRMDPs is similar. However, we apply Lipschitz conditions to the reward function rather than the classifier. A simple derivation shows that these interpretations are equivalent when the likelihood of the classifier is used to define the reward function.

\subsection{Related Work}

We aim to detect reward corruption by assuming the true reward to be ``smooth'' and the corruption to be ``spiky''. Smoothness of the reward function with respect to a distance in state space is a classic notion in machine learning \cite{Santamara1997ExperimentsWR}. Assumptions about the smoothness of the reward function have also been used to ensure safety properties for RL algorithms, for example by \cite{turchetta16safemdp} or \cite{Garcia2012}.

Both define smoothness with respect to a distance metric in the state space which is similar to our approach. However, they tackle the problem of safely exploring an MDP, i.e. without visiting dangerous states, and do not consider reward corruption. Another key difference are the assumptions on the distance functions, which are a subset of metrics that are connected to the (known) transition function in \cite{turchetta16safemdp} or are just the Euclidean distance in \cite{Garcia2012}. In contrast, we allow any metric.

There also exist approaches for automatically learning distance functions on the state space \cite{Taylor2011MetricLF,Globerson2005,Davis:2007:IML:1273496.1273523,NIPS2008_3446}. Such methods might be used in future work that remove the need for explicitly providing a distance function.

\section{Problem statement}

Let us recall the definition of CRMDPs from \cite{Everitt2017}.

\begin{definition}[CRMDP]
	A \emph{Corrupt Reward MDP} (CRMDP) is a finite-state MDP $\MDP$ with an additional \emph{corrupt reward function} $\corruption \colon \stateSpace \to \mathbb{R}$.
	We call $\MDP$ the \emph{underlying MDP}, $\stateSpace_n = \left\{ x \in \stateSpace \mid \rewardFunction(x) = \corruption(x) \right\}$ the set of \emph{non-corrupt states}, and its complement, $\stateSpace_c = \stateSpace \setminus \stateSpace_n$, the set of \emph{corrupt states}.
	\label{def:CRMDP}
\end{definition}

Note that $\corruption$ represents the reward observed by an agent and may be equal to the real reward $\rewardFunction$ in some, or even all, of the states. Our aim is to identify the corrupt states in $\stateSpace_c$ and learn an optimal policy with respect to $\rewardFunction$ while only observing $\corruption$. In general, this is impossible according to the CRMDP No Free Lunch Theorem. However special classes of CRMDPs may not have this limitation \cite{Everitt2017}. Therefore, we consider CRMDPs with a specific form of $\rewardFunction$ and $\corruption$.

\begin{definition}[Spiky CRMDP]
	Let $M$ be a CRMDP with two additional functions, $d$ and $\LV$.  $d \colon \stateSpace \times \stateSpace \to \mathbb{R}^{+}$ is a metric on the state space,
    and $\LV \colon \mathbf{Powerset}(\stateSpace) \times \stateSpace \to \mathbb{R}^{+}$ is non-decreasing with respect to set inclusion.
	We call $M$ a \emph{Spiky CRMDP} if the following assumptions are satisfied:

	\begin{enumerate}
        \item\label{cond:nonempty}$\stateSpace_n$ is nonempty
		\item \label{cond:smooth} $\forall{x, y \in \stateSpace}, ~ \absoluteValue{\rewardFunction(x) - \rewardFunction(y)} \leq d(x, y)$,
		\item \label{cond:spiky} $\forall {x \in \stateSpace_c}, ~ \LV_{\stateSpace_n}(x) > \sup_{y \in \stateSpace_n} \LV_{\stateSpace}(y)$.
	\end{enumerate}

    We call $d$ the \emph{distance} between the states, and $\LV$ the \emph{Lipschitz violation measure}.

	\label{def:spikyCRMDP}
\end{definition}
Intuitively, the distance $d$ should capture some notion of smoothness of
the true reward function in each state. The goal is to construct this distance such that the reward in corrupt states is much less smooth (hence,
``spiky''). The assumptions \ref{cond:smooth} and \ref{cond:spiky}
formalize this intuition.

Note the strong relationship between the distance and reward functions. For a given Spiky CRMDP one cannot be modified independently of the other without breaking the assumptions. In particular, any linear transformation applied to the reward, such as e.g. scaling, has to be also performed on the distance function.

The $\LV$ function is meant to be a measure of Lipschitz violations
of a state -- ``how much'' does a given state violate (\ref{cond:smooth}) with
respect to some set of states $\stateSpace_n$ when one substitutes
$\corruption$ for $\rewardFunction$.  We propose two ways of measuring this,
which further refine the class of Spiky CRMDPs. Unless otherwise noted, all our examples satisfy the conditions in definition \ref{def:spikyCRMDP} with either of these functions
used as $\LV$.

\begin{definition}[NLV]
	The \emph{Number of Lipschitz Violations} of $x \in \stateSpace$ with respect to $A \subseteq \stateSpace$ is
	\begin{equation*}
		\NLV_{A}(x) := \mu(\left\{ y \in A \mid \absoluteValue{C(x) - C(y)} > d(x, y) \right\}),
	\end{equation*}
	where $\mu$ is a measure on $\stateSpace$.
	\label{def:NLV}
\end{definition}
\begin{definition}[TLV]
	The \emph{Total Lipschitz Violation} of $x \in \stateSpace$ with respect to $A \subseteq \stateSpace$ is
	\begin{equation*}
		\TLV_{A}(x) = \int\displaylimits_{y \in A} \min\left\{ 0, \absoluteValue{C(x) - C(y)} - d(x, y) \right\} \textrm{ d} \mu(y),
	\end{equation*}
	where $\mu$ is a measure on $\stateSpace$.
	\label{def:TLV}
\end{definition}

Note that both variants require a measure on the state space.
In general there might not be a natural choice, but for finite state spaces, which we will consider, the counting measure is often reasonable. In this case, $\NLV$ counts the number of
states which violate the Lipschitz condition with the given state, while $\TLV$
sums up the magnitudes of the violations.

\section{Theoretical Results}

\subsection{Corruption identification}

With this setup in place, we can now introduce an algorithm to detect corrupt states in finite Spiky CRMDPs, which is shown in algorithm \ref{alg:identify}. The core idea is simple: We maintain a set of corrupt states, initially empty, and sort all states descending by their Lipschitz violation with respect to all states. Then for each state we check its Lipschitz violation with respect to all states that we have not identified as corrupt yet. If it is positive, we mark the state as corrupt. As soon as we encounter a state with zero Lipschitz violation we are done.

This algorithm makes use of assumptions \ref{cond:smooth} and \ref{cond:spiky} in definition \ref{def:spikyCRMDP}. Assumption \ref{cond:spiky} makes sure that by sorting the states in descending order we consider all corrupt states first. Assumption \ref{cond:smooth} then provides us with a simple stopping condition, namely no further violations of the Lipschitz condition.

\begin{algorithm}[bt]
	\begin{algorithmic}
		\Function{IdentifyCorruptStates}{$\stateSpace, \LV$} $\to \hat{\stateSpace}_c$
		\State $\hat{\stateSpace}_c \gets \emptyset$
		\State Sort $x \in \stateSpace$ by $\LV_\stateSpace(x)$ decreasing
		\For{$x \in \stateSpace$}
		\If{$\LV_{\stateSpace \setminus \hat{\stateSpace}_c}(x) = 0$}
		\State \Return $\hat{\stateSpace}_c$
		\EndIf
		\State Add $x$ to $\hat{\stateSpace}_c$
		\EndFor
		\EndFunction
	\end{algorithmic}
	\caption{An algorithm for identifying corrupt states in Spiky CRMDPs.}
	\label{alg:identify}
\end{algorithm}

\begin{proposition}
	Let $M$ be a spiky CRMDP. Then algorithm \ref{alg:identify}
	returns $\stateSpace_c$ when given $\stateSpace$ and $\LV$ as input.
	\label{prop:correctlyIdentifies}
\end{proposition}

This proposition simply states that the detection algorithm is able to correctly detect all corrupt states. The proof of this and all following statements is included in the appendix.

\subsection{A posteriori bounds on regret}

We now turn to the problem of learning an optimal policy despite the corruption in a CRMDP. Our first approach is to learn from an optimistic estimate of the true reward based on the Lipschitz condition on the rewards. To this end we first define such upper and lower bounds on the rewards.

\begin{definition}
	Let $M = \SCRMDP$ be a spiky CRMDP. Then for a state $x$ we define the \emph{reward lower Lipschitz bound} to be
	\begin{equation*}
        \rllb(x) = \max_{y \in \stateSpace_n} \rewardFunction(y) - d(x, y).
	\end{equation*}
    We call any state that introduces this bound and is closest to $x$ the
    \emph{lower Lipschitz bounding state}.  Symmetrically we define the upper
    bounds and upper bounding states.
	\begin{equation*}
		\rulb(x) = \min_{y \in \stateSpace_n} \rewardFunction(y) + d(x, y).
	\end{equation*}
	\label{def:bounds}
\end{definition}

For a non-corrupt state, both bounds just equal the true reward, so it's its own bounding state.
We also get $\rllb(x) \leq \rewardFunction(x)$ and $\rulb(x) \geq \rewardFunction(x)$,
because the distance function is positive definite.

Note that the reward lower (or upper) Lipschitz bound and bounding state can be
computed by the agent after identifying the corrupt states, because it only
requires access to the distance function and real reward function for
non-corrupt states, which is equal to the corrupt reward there.

After computing these bounds, the agent can compute upper bounds on the regret
it is experiencing. Finding a policy with respect to this optimistic estimate of the true reward function of corrupt states gives us a way to bound the expected regret using the Lipschitz bounds.

\begin{proposition}
	The expected regret with respect to $R$ of a policy $\pi'$ optimal with respect to
	$\rulb$ (the reward \emph{upper} Lipschitz bound) is bounded from above by
	\begin{equation}
		\sup_{\pi \textrm{ $\rulb$-optimal}} \E_{\tau \sim \pi} \sum_{x \in \tau} \rulb(x) - \rllb(x).
		\label{eq:regretBound}
	\end{equation}
	\label{prop:regretBound}
\end{proposition}

This bound is not particularly useful as a theoretical result, because it is
fairly easy to contrive CRMDPs where it becomes as large as the
difference between the least and greatest possible cumulative rewards.

However, it might be useful to increase sample efficiency in an active reward learning setting.
Say we have a supervisor that we can ask to provide us with the real reward of
a given state, but such a question is expensive.  We therefore want to maximize
the information we get from a single question.  The way we compute the bound
\eqref{eq:regretBound} allows us to pick a question such that the upper bound
on regret improves the most, which is a good criterion for question quality. We do not investigate this further, but suggest it as useful future work.

\subsection{Optimality with corruption avoidance}

To be able to guarantee an optimal policy despite corruption, we have to make an additional assumption about the environment. In particular, we will assume that the underlying MDP has at least one optimal policy which avoids all corrupt states. This essentially means that identifying and then avoiding the corrupt states is enough to solve the environment.

\begin{proposition}
    Let $M = \SCRMDP$ be a spiky reward CRMDP and $\dot{M} = \MDP$ its
    underlying MDP. Then if there exists a $\dot{M}$-optimal policy $\pi^*$
    generating a trajectory $\tau \sim \pi^*$ that does not contain any corrupt states, then
    any policy optimal with respect to $M$ using $\rllb$ as a reward function will also be $\dot{M}$-optimal.
	\label{prop:optimalCorruptionAvoidance}
\end{proposition}

The assumption of an optimal policy that always avoids corrupt states is very
strong, especially in stochastic environments.  However, this is to be expected
since our result allows for solutions without any regret.

It is also worth noting that the assumption might be slightly weakened in practice, as we discuss in our experimental results.

\section{Practical considerations}

\begin{algorithm}[bt]
	\begin{algorithmic}
		\Function{LearnOnline}{$\pi, \LV$}
		\State $\hat{\stateSpace}_c \gets \emptyset$
		\For{$\tau \sim \pi$}
		\State $X \gets IdentifyCorruptStates(\tau, \LV)$
		\State $\hat{\stateSpace}_c \gets \hat{\stateSpace}_c \cup X$
		\State $\hat{\stateSpace_n} \gets \hat{\stateSpace_n} \cup (\tau \setminus \hat{\stateSpace}_c)$
		\State Train RL agent using reward signal $\rllb_{\hat{\stateSpace}_n}$
		\EndFor
		\EndFunction
	\end{algorithmic}
	\caption{An example of using algorithm \ref{alg:identify} online when \ref{cond:locallySpiky} is satisfied.}
	\label{alg:learnOnline}
\end{algorithm}

Algorithm \ref{alg:identify} sorts over an entire state space, which requires (1) complete knowledge of the state space and (2) computational resources to perform a sorting operation. 
\subsection{Modification for Online Learning}
We would like our algorithm to work online, i.e. at most considering a small batch of trajectories at once. Our current
assumptions are not enough for such an algorithm to be correct. However, it is possible by strengthening assumption \ref{cond:spiky} in definition \ref{def:spikyCRMDP}:
\begin{enumerate}[label=\textnormal{\arabic*'}]
	\setcounter{enumi}{2}
	\item \label{cond:locallySpiky} $\forall {\pi, \tau \sim \pi} ~ \forall {x \in \tau_c}, ~ \LV_{\tau_n}(x) > \sup_{y \in \tau_n} \LV_{\tau}(y)$,
\end{enumerate}
where $\tau \sim \pi$ is a trajectory sampled from policy $\pi$. We treat the trajectory $\tau$ as a sequence of states, $\tau_c$ are
the corrupt states in this sequence, and $\tau_n = \tau \setminus \tau_c$.

This is the same condition as before, except that we require it to hold over all possible trajectories through the MDP. Functionally speaking, this allows us to be sure that we'll be able to iteratively perform corruption identification without misidentifying any states. This assumption is much stronger than the previous version and restricts the
class of environments satisfying it considerably. In practice, we believe that
many interesting environments will still satisfy it, and many of those that do not may still be learnable in a similar manner.

With assumption \ref{cond:locallySpiky} satisfied, we can use
algorithm \ref{alg:identify} for online reinforcement learning, as shown in algorithm \ref{alg:learnOnline}. To do this, we sample trajectories from the current policy of the agent, apply the algorithm \ref{alg:identify} on the individual trajectories and then update the policy using the reward lower Lipschitz bound.

Note that $\rllb_{\hat{\stateSpace}_n}(x)=\max_{y \in \hat{\stateSpace}_n} \rewardFunction(y) - d(x, y)$.  A straightforward application of proposition \ref{prop:correctlyIdentifies} shows that $\hat{\stateSpace}_c \rightarrow \stateSpace_c$ from below and $\hat{\stateSpace}_n \rightarrow \stateSpace_n$ from above when using algorithm \ref{alg:learnOnline}.

In order to actually use the identified states, we also need
the ability to compute the $\rllb$ and $\rulb$ functions. This once again
would normally require access to the whole state space with corrupt states identified. However, we can approximate $\rllb$ and $\rulb$ by slowly building up the known state space and
corrupt state space. This is what is reflected in algorithm \ref{alg:learnOnline}, specifically when using $\rllb_{\hat{\stateSpace}_n} \approx \rllb$ as a reward signal. This approximation is pessimistic for corrupt states but converges as $\hat{\stateSpace}_n \rightarrow \stateSpace_n$.

\subsection{Memory complexity}

Memory complexity is an additional challenge for our algorithm. Keeping the whole state space in memory is usually not feasible, and even keeping only the encountered states can quickly result in performance problems. We implemented some optimizations to reduce the memory consumption of our approach. While they had no effect in the small toy environments we consider in this paper, for completeness and future reference we include a description of these optimizations in appendix \ref{app:memory}.

\section{Experiments}\label{sec:experiments}
\begin{table*}[t]
    \centering
    \begin{tabular}{|l|l|l|c|c|c|c|}
     \hline
     Environment               & Reward    & Agent    & Avg. Corrupt Reward & Avg. True Reward & Sample Complexity & SC ratio \\
     \hline
     \multirow{4}{*}{Corners}  & Corrupt   & Baseline & 73             & 48            &              5421 & 0.17     \\
                               & Uncorrupt & Baseline & 64             & 64            &             31410 & 1.00     \\
                               & Uncorrupt & CRMDP    & 64             & 64            &             32250 & 1.03     \\
                               & Corrupt   & CRMDP    & 64             & 64            &             57000 & 1.81     \\
     \hline
     \multirow{4}{*}{OnTheWay} & Corrupt   & Baseline & 73             & 48            &              4194 & 0.09     \\
                               & Uncorrupt & Baseline & 64             & 64            &             45310 & 1.00     \\
                               & Uncorrupt & CRMDP    & 64             & 64            &             51750 & 1.14     \\
                               & Corrupt   & CRMDP    & 64             & 64            &             94380 & 2.08     \\
     \hline
    \end{tabular}
    \caption{Results of the gridworld experiments described in section \ref{sec:experiments}. In addition to the final corrupted and hidden true reward we report the \emph{sample complexity} of each, which is defined as the number of episodes required for a moving average (momentum=0.9) of observed return to reach its optimal value. The \emph{SC ratio} is the ratio of the sample complexity compared to the baseline model that has access to the hidden true reward function. Note that sample complexity measures are generally susceptible to noise, and should be interpreted with caution.}
    \label{tab:results}
\end{table*}

We ran experiments on gridworld environments, detailed below. On each environment, we trained three different agents using an implementation of PPO \cite{Schulman17}.

The first agent just uses PPO with access to the corrupt reward, without any consideration for corruption.

The second agent uses PPO with access to the hidden reward, with the corruption removed. These are two baselines -- how well an algorithm performs on the corrupt state and how well it could perform on the environment if it was not corrupted.

The third agent uses algorithm \ref{alg:learnOnline} during the rollout phase of the PPO algorithm. In particular, it identifies corrupt states in the rollouts and replaces their rewards by the Lipschitz bounds. The full code used, logs and commands to reproduce the experiments can be found at: \url{https://github.com/jvmancuso/safe-grid-agents}.

To evaluate the quality of our algorithm for each experiment we calculate the average corrupt reward, average hidden reward, the sample complexity needed to achieve this result, and the ratio of this complexity to the complexity needed in the non-corrupt baseline.
The results are summarized in Table \ref{tab:results} and we proceed by discussing them in detail.

\subsection{Toy Spiky-CRMDP}

Similar to \cite{Everitt2017}, we construct a toy example under
which our learnability guarantee from proposition
\ref{prop:optimalCorruptionAvoidance} is satisfied. We also slightly tweak this
toy example to break the requirements for proposition
\ref{prop:optimalCorruptionAvoidance}, with the hope of demonstrating that the
theorem's requirements can be relaxed to some extent without harming learnability. The figure
shows the toy example and its modification.

\begin{figure}[t]
    \centering
    \input{toy-spiky-crmdp.tex}
    \caption{Toy Spiky CRMDP envorinments \emph{Corners} on the left, \emph{OnTheWay} on the right. The blue cell in the lower right hand corner is the starting position of the agent and the green cell in the upper left hand corner the goal. The true reward collected in each state is determined by the max-distance to the goal and shown here by the numbers in each cell. The reward in the red cells is corrupted and the underlying true reward is shown in parentheses.}
    \label{fig:toy}
\end{figure}
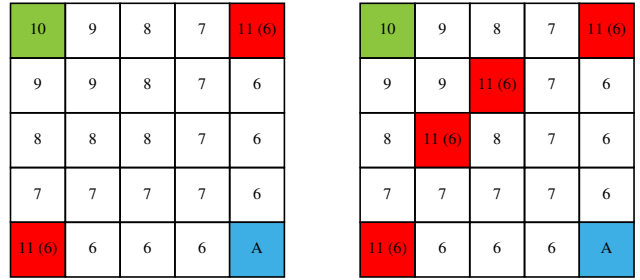

The gridworlds are shown in \ref{fig:toy}. The agent starts on the blue field and, in typical gridworld fashion, can move up/down/left/right as its action. The red cells are corrupted states and give the agent an unusually high reward, thereby satisfying our conditions about the ``spikyness'' of the corruption. We use the Manhattan metric as our distance measure. Note that in the environment on the right, the optimal policy must encounter corrupt states, violating the assumption of proposition \ref{prop:optimalCorruptionAvoidance}. However, because the optimal policy remains optimal after substituting $\rllb$ for the reward we should still expect good performance from our algorithm.

\subsection{Results}

The baseline results for both environments are as expected. PPO with access to the corrupt reward very quickly learns the corrupt-optimal policy, going straight to the bottom left or top right corner. PPO with access to the hidden reward needs significantly more data to learn the optimal policy, because the problem is more complicated and cannot be solved with a constant policy.

As a sanity check we ran an additional baseline test for these toy environments -- our algorithm with access to the hidden reward. As it does not encounter any states it could perceive as corrupt it performs comparably to the baseline.

Finally we ran our algorithm without access to the hidden reward. In both cases it learned the optimal policy, requiring about two times as much data as the agent with access to the hidden reward. It is worth pointing out that because of the way the environments are constructed, this additional data is most likely \emph{not} used for learning the bounds on the true reward. These bounds will almost always be as good as they can as soon as the agent identifies the corruption. Rather, the increased difficulty probably stems from the lower differences between optimal and sub-optimal policy payoffs.

\section{Conclusion}
 The class of Spiky CRMDPs resolves several limitations in the class of previously known, solvable CRMDPs. In particular, this class of MDPs need not have finite diameter (state spaces symmetric in time), we demonstrate that they can be solved in the usual MDP formalism without recourse to the decoupled RL of \cite{Everitt2017}.

Despite the the experimental support for our algorithm's success in toy gridworld environments, there are several limitations of our solution. Even though we can minimize the regret, it required quite a few assumptions to do so. Our results for the \emph{OnTheWay} environment suggest that these assumptions can be weakened further, and this could be useful future work.  However, we believe the regret bound is most intriguing, as it can be used for accelerating exploration in decoupled RL schemes. This is most apparent for semi-supervised RL, but also applies to other settings in which reward information can be inferred from external channels or actors.  For this bound to become practically useful, future work should prioritize learning the Spiky CRMDP distance metric $d$ from trajectory data in an online or active reward learning setting.

More generally, Lipschitz reward functions and spiky corruption can be seen as a particularly strong prior in the Bayesian RL setting. Our theorems and experiments demonstrate that this can be used to encode useful inductive biases in relevant environments. While we demonstrate a single instance of its usefulness in learning within misspecified environments, these priors can be very useful in a wide variety of practical settings in which Bayesian RL has traditionally fallen short.

\section{Acknowledgements}

Thanks to Victoria Krakovna and Tom Everitt for generous feedback and insightful discussions about this work and to Rohin Shah, Jan Leike, and Geoffrey Irving for providing valuable feedback on early proposals of our approach.

Furthermore, thanks to the organizers of the 2nd AI Safety Camp during which a significant portion of this work was completed.

\bibliographystyle{named}
\bibliography{ijcai19}

\appendix

\section{Proofs of theoretical results}

Proof of proposition \ref{prop:correctlyIdentifies}:
	\begin{proof}
		First note that
		\begin{equation*}
			\forall_{x \in \stateSpace_c} \LV_{\stateSpace}(x) \geq \LV_{\stateSpace_n}(x) > \sup_{y \in \stateSpace_n} \LV_{\stateSpace}(y),
		\end{equation*}
		because in general $\LV$ is non-decreasing with respect to inclusion, that is $\LV_A \geq \LV_B$ if $A \subseteq B$.
		This immediately tells us that all corrupt states $x \in \stateSpace_c$ are processed by Algorithm \ref{alg:identify} before all
		non-corrupt states $y \in \stateSpace_n$.

		We will show that all states added to the $\hat{\stateSpace}_c$ set are actually corrupt.
		For a state $x \in \stateSpace$ to be added to that set it has to satisfy
		\begin{equation*}
			\LV_{\stateSpace \setminus \hat{\stateSpace}_c} (x) > 0.
		\end{equation*}
		This condition can only be satisfied only if $\stateSpace_c \setminus \hat{\stateSpace}_c \neq \emptyset$, because otherwise
		$\stateSpace \setminus \hat{\stateSpace}_c$ would contain no corrupt states and no non-corrupt states can violate the Lipschitz condition.
		But Algorithm \ref{alg:identify} processes all corrupt states before any non-corrupt states, so $x$ has to be among the corrupt ones.

        The algorithm returns as soon as it finds a state $x$ satisfying
        $\LV_{\stateSpace \setminus \hat{\stateSpace}_c}(x) = 0$.
        Because of the processing order it is sufficient to prove that such a
        state is a non-corrupt state.  Note that
		\begin{equation*}
			\LV_{\stateSpace \setminus \hat{\stateSpace}_c}(x) \geq \LV_{\stateSpace_n}(x),
		\end{equation*}
		because, as we have already shown, $\hat{\stateSpace}_c \subseteq \stateSpace_c$.
        Since the left hand side of this expression is zero and the right one
        is nonnegative by definition, it also has to be zero.  But zero cannot
        be strictly greater than a nonempty supremum of nonnegative values, so
        $x$ cannot be corrupt.
	\end{proof}
	
Proof of proposition \ref{prop:regretBound}:
	\begin{proof}
		First note that any policy $\pi$ has average cumulative reward with respect to $\rulb$ greater or equal than as with respect to $R$
		\begin{equation}
			\E_{\tau \sim \pi} \sum_{x \in \tau} \rewardFunction(x) \leq \E_{\tau \sim \pi} \sum_{x \in \tau} \rulb(x),
			\label{eq:lowerIsLess}
		\end{equation}
		because in general $\rulb(x) \geq \rewardFunction(x)$ as the only difference between these functions is a substitution of upper bounds for some, possibly lower, values of $R$.
		In particular, since this is also true for $R$-optimal policies, this means that any $\rulb$-optimal policy $\pi'$ will have average cumulative reward
		with respect to $\rulb$ greater or equal than the $R$-optimal policy $\pi^*$ has with respect to $R$:
		\begin{equation}
			\E_{\tau \sim \pi^*} \sum_{x \in \tau} \rewardFunction(x) \leq \E_{\tau \sim \pi^*} \sum_{x \in \tau} \rulb(x) \leq \E_{\tau \sim \pi'} \sum_{x \in \tau} \rulb(x).
			\label{eq:lessRealThanUpper}
		\end{equation}

		Now, by \eqref{eq:lowerIsLess}, we get another bound on average cumulative reward of any policy $\pi$
		\begin{equation*}\begin{split}
			\E_{\tau \sim \pi} \sum_{x \in \tau} \rulb(x) + (\rllb(x) - \rulb(x)) &= \E_{\tau \sim \pi} \sum_{x \in \tau} \rllb(x)\\ &\leq \E_{\tau \sim \pi} \sum_{x \in \tau} \rewardFunction(x).
		\end{split}\end{equation*}
		By moving the part in parentheses to the right side of the inequality we get
		\begin{equation}
			\E_{\tau \sim \pi} \sum_{x \in \tau} \rulb(x) \leq \E_{\tau \sim \pi} \sum_{x \in \tau} \rewardFunction(x) + (\rulb(x) - \rllb(x)).
			\label{eq:regretInSpecificTrajectory}
		\end{equation}
		Using \eqref{eq:lessRealThanUpper} and \eqref{eq:regretInSpecificTrajectory} for $\rulb$-optimal policies $\pi'$ we get
		\begin{equation*}\begin{split}
			\E_{\tau \sim \pi^*} \sum_{x \in \tau} \rewardFunction(x) &\leq \E_{\tau \sim \pi'} \sum_{x \in \tau} \rulb(x)\\ &\leq \E_{\tau \sim \pi'} \sum_{x \in \tau} \rewardFunction(x) + (\rulb(x) - \rllb(x)).
		\end{split}\end{equation*}
		We get the final bound by moving the reward to the left hand side and taking a supremum over $\rulb$-optimal policies of both sides.
		\begin{multline*}
			\sup_{\pi' \textrm{ $\rulb$-optimal}} \E_{\tau \sim \pi^*} \sum_{x \in \tau} \rewardFunction(x) - \E_{\tau \sim \pi'} \sum_{x \in \tau} \rewardFunction(x) \leq\\ \sup_{\pi' \textrm{ $\rulb$-optimal}} \E_{\tau \sim \pi'} \sum_{x \in \tau} (\rulb(x) - \rllb(x)).
		\end{multline*}
	\end{proof}

Proof of proposition \ref{prop:optimalCorruptionAvoidance}:
	\begin{proof}
        Recall the inequality in \eqref{eq:lowerIsLess}, that is
		\begin{equation*}
			\E_{\tau \sim \pi} \sum_{x \in \tau} \rllb(x) \leq \E_{\tau \sim \pi} \sum_{x \in \tau} \rewardFunction(x).
		\end{equation*}
        This means that $\pi^*$ is $\rllb$-optimal, because its average cumulative reward
        does not change, so it is still the greatest possible.  Any other
        $\rllb$-optimal policy has to get at least as much average cumulative
        reward as $\pi^*$ with respect to $\rllb$.  Since it would get at least
        as much reward with respect to $R$, it is also $R$-optimal.
	\end{proof}
	
\section{Reducing memory consumption}\label{app:memory}

We cannot avoid to store all the corrupt states, as we need to substitute our approximations for the rewards received when encountering them. However, keeping all the non-corrupt states in memory is not strictly necessary, because they are only used to improve our approximation of $\rllb$. To reduce memory consumption, we can instead keep only a small set of them and use it to update the cached $\rllb$ of all known corrupt states. We add newly encountered non-corrupt states to this set, but if it gets too large, we remove some states at random. 

Since we always update our approximation of $\rllb$ using newly encountered non-corrupt states, it converges to the correct value as long as a state giving the best bound is not encountered earlier than the corrupt state. Because of the stronger version of assumption \ref{cond:spiky}, we can even expect the state giving the best bound to be in the very trajectory identifying the corrupt state. Because of this we do not expect any practical problems with this optimization. All of our experiments use this modification of algorithm \ref{alg:learnOnline}; however, in the toy environments presented the size of the cached set was bigger than the state space, thus there was no practical effect.

Future work could further reduce memory consumption by keeping the information about the state space in different ways, for example using neural networks to approximate the required functions.

\end{document}

%% file: toy-spiky-crmdp.tex
\resizebox{0.45\columnwidth}{!}{%
\begin{tikzpicture}[line width=1pt]
  \matrix (m)[matrixstyle]
  {
    \node[fill=LimeGreen]{10}; \& 9 \& 8 \& 7 \& \node[fill=red]{11 (6)}; \\
    9 \& 9 \& 8 \& 7 \& 6 \\
    8 \& 8 \& 8 \& 7 \& 6 \\
    7 \& 7 \& 7 \& 7 \& 6 \\
    \node[fill=red]{11 (6)}; \& 6 \& 6 \& 6 \& \node[fill=CornflowerBlue]{A}; \\
  };
\end{tikzpicture}
}
\qquad
\resizebox{0.45\columnwidth}{!}{%
\begin{tikzpicture}[line width=1pt]
  \matrix (m)[matrixstyle]
  {
    \node[fill=LimeGreen]{10}; \& 9 \& 8 \& 7 \& \node[fill=red]{11 (6)}; \\
    9 \& 9 \& \node[fill=red]{11 (6)}; \& 7 \& 6 \\
    8 \& \node[fill=red]{11 (6)}; \& 8 \& 7 \& 6 \\
    7 \& 7 \& 7 \& 7 \& 6 \\
    \node[fill=red]{11 (6)}; \& 6 \& 6 \& 6 \& \node[fill=CornflowerBlue]{A}; \\
  };
\end{tikzpicture}
}